\documentclass{article}
\usepackage{spconf,amsmath,graphicx,xcolor,comment,hyperref,amssymb}

\title{A FEW-SHOT ATTENTION RECURRENT RESIDUAL U-NET FOR CRACK SEGMENTATION}

\name{\begin{tabular}{c}Iason Katsamenis$^{\star}$, Eftychios Protopapadakis$^{\dagger}$, Nikolaos Bakalos$^{\star}$, \\ Anastasios Doulamis$^{\star}$, Nikolaos Doulamis$^{\star}$ and Athanasios Voulodimos$^{\star}$\end{tabular} \thanks{This work has received funding from the European Union’s Horizon 2020 Research and Innovation Programme under grant agreement No 955356 (Improved Robotic Platform to perform Maintenance and Upgrading Roadworks: The HERON Approach).}}

\address{$^{\star}$ National Technical University of Athens, 9\textsuperscript{th} Iroon Polytechniou str., 15773 Athens, Greece \\ $^{\dagger}$ University of Macedonia, 156\textsuperscript{th} Egnatia str., 54636 Thessaloniki, Greece\\
}

\begin{document}


%
\maketitle
\begin{abstract}

Recent studies indicate that deep learning plays a crucial role in the automated visual inspection of road infrastructures. However, current learning schemes are static, implying no dynamic adaptation to users’ feedback. To address this drawback, we present a few-shot learning paradigm for the automated segmentation of road cracks, which is based on a U-Net architecture with recurrent residual and attention modules (R2AU-Net). The retraining strategy dynamically fine-tunes the weights of the U-Net as a few new rectified samples are being fed into the classifier. Extensive experiments show 
that the proposed few-shot R2AU-Net framework outperforms
other state-of-the-art networks in terms of Dice and IoU metrics, on a new dataset, named CrackMap, which is made publicly available at \href{https://github.com/ikatsamenis/CrackMap}{https://github.com/ikatsamenis/CrackMap}.
\end{abstract}
\section{Introduction}
\label{sec:intro}

The development of cracks on the road surface is a frequently occurring defect and can constitute a safety hazard for road users. Cracking in its various types (longitudinal, oblique, alligator cracks, etc.) affects the traffic flow and safety, resulting in poor performance of the road infrastructure, accidents, as well as increased $CO_2$ emissions, fuel costs, and time delays. Indicatively, for 2006, the comprehensive cost of traffic crashes where road conditions contributed to crash occurrence or severity, in the United States alone, is estimated at \$217.5 billion, which corresponds to 43.6\% of the total crash costs \cite{zaloshnja2009cost}. More recent evidence highlights that approximately \$400 billion is invested globally each year in pavement construction and maintenance \cite{torres2015sustainable}. Therefore, the adoption of effective monitoring strategies can lead to enormous economic, social, and environmental benefits to the community.

Recently, there is a great research interest in the automatic visual inspection of road distress, by analyzing visual data. Generally, deteriorated pavement produces rough surfaces, which entails that various image processing methods such as thresholding \cite{7077805}, edge detection \cite{5646923}, and mathematical morphology \cite{tanaka1998crack} can be used to localize crack regions. The core idea behind these approaches is that cracked regions tend to demonstrate non-uniformity, while on the other hand, the color and textural characteristics of the non-deteriorated road surface are more consistent and smoother. However, even though such techniques are computationally efficient, they are susceptible to image noise and fail to generalize the differentiation between the defect and the surface background \cite{9121269}.

Current developments in deep learning and artificial intelligence technology have led Convolutional Neural Networks (CNNs) to be an effective tool for the automatic visual inspection of road infrastructures \cite{9038607, 9081001, pandey2022convolution}. The main asset of such deep architectures, compared to the aforementioned conventional image processing methods, is the fact that they leverage throughout the learning procedure annotated ground truth data \cite{KATSAMENIS2022104182}. Thereby, these algorithms demonstrate high identification accuracy by effectively learning the essential features needed to classify a given pixel as defective or not.

Usually, Fully Convolutional Networks (FCNs), or their variants U-Nets, are considered for providing a precise pixel-based segmentation of damaged areas from RGB images of road infrastructures \cite{9121269, long2015fully, 10.1007/978-3-319-24574-4_28}. This is mainly due to the fact that FCNs have emerged as powerful segmentation tools, especially for performing accurate pixel-based classification tasks for challenging problems (e.g., biomedical imaging problems with data expressed either in 2D or 3D \cite{9053405, s21062215}, as well as crack segmentation \cite{jenkins2018deep}). To enhance the performance of the original U-Net, numerous elements have been introduced, such as residual convolutional units \cite{8309343, alom2018recurrent} and attention gates instead of the typical skip connection \cite{oktay2018attention, 8803060}.


%
%

\begin{figure*}[htb]

\begin{minipage}[b]{.6\linewidth}
  \centering
  \centerline{\includegraphics[width=10.35cm]{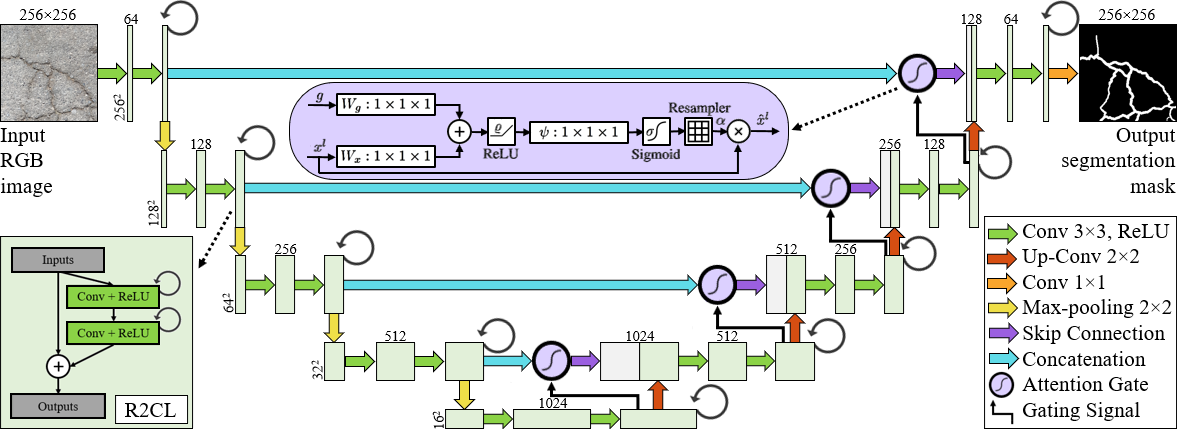}}
  \small\centerline{(a) The proposed U-Net architecture with recurrent residual and attention modules.}
\end{minipage}
\hfill
\begin{minipage}[b]{0.38\linewidth}
  \centering
  \centerline{\includegraphics[width=6.6cm]{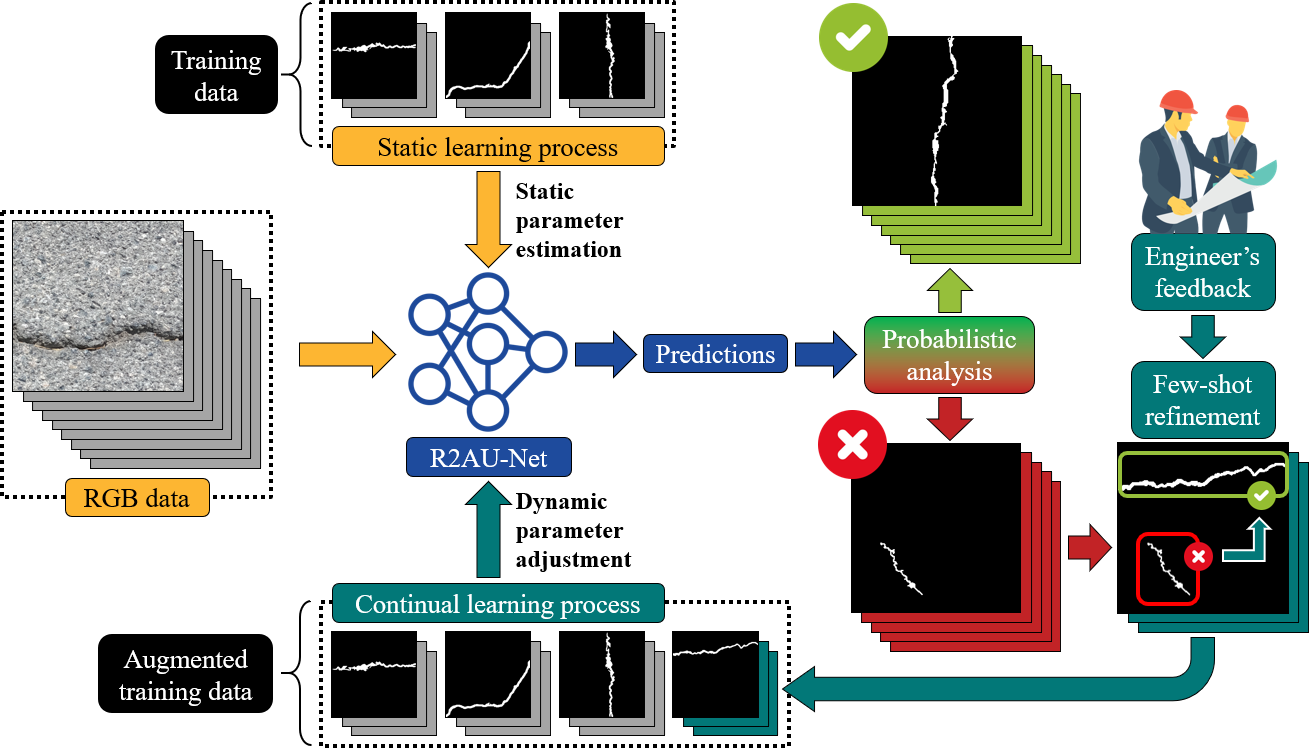}}
  \small\centerline{(b) The proposed few-shot retraining strategy.}
\end{minipage}
\caption{A schematic representation of the proposed (a) R2AU-Net and (b) few-shot learning scheme for road crack segmentation.}
\label{fig:architecture}
\end{figure*}

\subsection{Current limitations and our contribution}



Inspired by the above research work, we present R2AU-Net, which is a deep U-Net structure with recurrent residual and attention modules (see Fig. \ref{fig:architecture}a). Compared to the standard U-Net, R2AU-Net incorporates recurrent residual convolutional layers (R2CL) that ensure better feature representation for the segmentation task and attention gates to highlight salient features which are passed through the skip connections \cite{alom2018recurrent, oktay2018attention}.

Still, however, the most crucial hindrance of the aforementioned typical deep learning frameworks is that they treat the segmentation task as a static procedure. More specifically, they leverage knowledge derived from labeled data, but, nevertheless, it is not possible to further refine their outputs by exploiting user's interaction, especially in cases where the deep network underperforms \cite{s21062215}. To this end, we introduce a few-shot refinement scheme which is a semi-supervised learning paradigm, based on the R2AU-Net, that is able to adapt the model's behavior and weights according to user’s feedback, to further increase the segmentation performance (see Fig. \ref{fig:architecture}b).

\section{Proposed Architecture}
\label{sec:proposed}

\subsection{R2AU-Net architecture for road crack segmentation}

In this section, we present R2AU-Net, whose architecture is illustrated in Fig. \ref{fig:architecture}a and is a combination of Recurrent Residual U-Net \cite{alom2018recurrent} and Attention U-Net \cite{oktay2018attention}. It is emphasized that the model was designed for segmenting cracks in RGB images  and, hence, provides meticulous information on a variety of metrics and properties that are critical for the automated and robotic-driven maintenance process, such as geometry, type, orientation, length, density, and shape of cracks.

In particular, the operations within the Recurrent Convolutional Layers (RCL) in R2CL (see Fig. \ref{fig:architecture}a) are carried out based on the discrete time steps which are expressed according to the RCNN \cite{liang2015recurrent}. Suppose we have an input at layer $l$ within an R2CL block, and a pixel located at $(i,j)$ within an input on the feature map $k$ in the RCL. We denote $\mathcal{Y}^l_{ijk}(t)$ the output of the model at time step $t$, which can be expressed as:

\vspace{-5mm}
\begin{equation}
\mathcal{Y}^l_{ijk}(t)=(w_k^\varrho)^T \cdot x_l^{\varrho(i,j)}(t) + (w_k^r)^T \cdot x_l^{r(i,j)}(t-1) + \beta_k
\label{eq:recres1}
\end{equation}
\vspace{-5mm}

\noindent
where $x_l^{\varrho(i,j)}(t)$ and $x_l^{r(i,j)}(t-1)$ represent respectively the inputs to the standard convolution layers and $l^{th}$ RCL. In parallel, $w_k^\varrho$ and $w_k^r$ denote respectively the weights of the standard convolutional layer and RCL that correspond to the $k^{th}$ feature map, whereas $\beta_k$ symbolizes the bias. The RCL's output is activated by the ReLU function $\varrho$ as follows:

\vspace{-5mm}
\begin{equation}
\mathcal{R}(x_l,w_l)=\varrho(\mathcal{Y}^l_{ijk}(t))=max(0,\mathcal{Y}^l_{ijk}(t))
\label{eq:recres2}    
\end{equation}
\vspace{-5mm}

Let $x_l$ an input sample of the R2CL unit, then, the R2CL's output $x_{l+1}$, in both the downsampling layer in the encoding path and the upsampling layer in the decoding path of the R2AU-Net, can be calculated using the following equation:

\vspace{-4mm}
\begin{equation}
x_{l+1}=x_l+\mathcal{R}(x_l,w_l)
\label{eq:recres3}    
\end{equation}
\vspace{-5mm}

In parallel, as one can observe in Fig. \ref{fig:architecture}a we integrate into R2AU-Net an attention gate mechanism to focus on points and shapes of interest (i.e., road cracks) \cite{oktay2018attention}. More specifically, for each pixel $i$, the attention coefficients $\alpha_i \in [0,1]$ tend to yield higher values in target crack regions and lower values in background road areas. We obtain the output of the attention gate in layer $l$ by multiplying element-wise the input feature maps and attention coefficients: $\hat{x}_i^l=x_i^l \cdot \alpha_i^l$. Attention values are calculated for each pixel vector $x_i^l \in \mathbb{R}^{F_l}$, where $F_l$ denotes the number of feature maps in layer $l$. Also, a gating vector $g_i \in \mathbb{R}^{F_g}$ is utilized in order to determine the focus area per pixel. To achieve greater performance the attention coefficient is derived by leveraging the additive attention:

\vspace{-5mm}
\begin{equation}
Q^l_\alpha = \psi^T(\varrho(W_x^Tx_i^l + W_g^Tg_i + \beta_g)) + \beta_\psi \textrm{,} \quad a_i^l = \sigma(Q^l_\alpha)
\label{eq:att1}    
\end{equation}
\vspace{-5mm}

\noindent
where $\sigma$ corresponds to the sigmoid activation function, $W_x \in \mathbb{R}^{F_l\times F_{int}}$ and $W_g \in \mathbb{R}^{F_g\times F_{int}}$ are linear transformations that are calculated by utilizing channel-wise 1×1×1 convolutions for the input tensors, and, lastly, $\beta_g \in \mathbb{R}^{F_{int}}$ and $\beta_\psi \in \mathbb{R}$ denote the bias.

\subsection{Few-shot learning for segmentation refinement}
\label{sec:online}

As shown in Fig. \ref{fig:architecture}b, we hereby propose a dynamic rectification scheme that leverages expert users’ feedback on a small part of the data in order to improve the overall performance of the aforementioned R2AU-Net. The proposed retraining strategy dynamically updates the weights of the model, so that (a) the refined incoming samples are trusted as much as possible, while simultaneously (b) a minimal degradation of the already gained knowledge is achieved. To this end, let us denote $p_{ij}$ the soft label value of a pixel that is located in position $(i,j)$ of a given image. Then, for each input $n$ we calculate the average image confidence score $I_n$, defined as:

\vspace{-3mm}
\begin{equation}
I_n = \frac{1}{\sum_{\forall i,j}{\zeta_{ij}}} \cdot \sum_{i=1}^{R} \sum_{j=1}^{C} \zeta_{ij}\cdot p_{ij}
\label{eq:cofnscore}    
\end{equation}

\noindent
where C and R correspond to the image's columns and rows respectively. In parallel, $\zeta_{ij} \in \{0,1\}$ equals 1 when $p_{ij} > \vartheta$ and 0 otherwise, where $\vartheta$ is the detection acceptance threshold, which is set to 0.5. As such, the confidence score considers only the cracked regions over the image $n$, provided by the deep classifier. Subsequently, we rank the images according to $I_n$ scores. The 5\% of the lower ranked images are provided to an engineer expert, who rectifies the model's segmentation outputs. Lastly, the refined few-shot annotated data are fed back to the network for updating the model's weights.

\section{EXPERIMENTAL SETUP AND RESULTS}
\label{sec:experimental}

\subsection{Dataset description}
\label{sec:data}

For the training procedure five datasets, consisting of 4,717 images in total, that depict crack defects, were utilized (see Table \ref{table:data}). 
During the data preprocessing step, the RGB data were resized to a resolution of 256×256 pixels. Lastly, 80\% of the data was used for training the models (3,774 images), while the rest 20\% was used for validation (943 images).

\begin{table}[h!]
\small
\centering
\begin{tabular}{||c c c||} 
 \hline
 \textbf{Set} & \textbf{Name} & \textbf{Number of RGB samples} \\ [0.5ex] 
 \hline\hline
   & CFD & 118 \\ 
   & CRACK500 & 3,363 \\
 \textbf{Train and} & Cracktree200 & 206 \\
 \textbf{Validation} & DeepCrack & 521 \\
   & GAPS384 & 509 \\
   & \textbf{Total} & \textbf{4,717} \\  [0.5ex] 
 \hline\hline
 \textbf{Test} & \textbf{CrackMap} & \textbf{120 -- 6} \\
 \hline
\end{tabular}
\caption{Utilized datasets for training and evaluation tasks.}
\label{table:data}
\end{table}

\begin{figure}[htb]

\begin{minipage}[b]{.48\linewidth}
  \centering
  \centerline{\includegraphics[width=4.0cm]{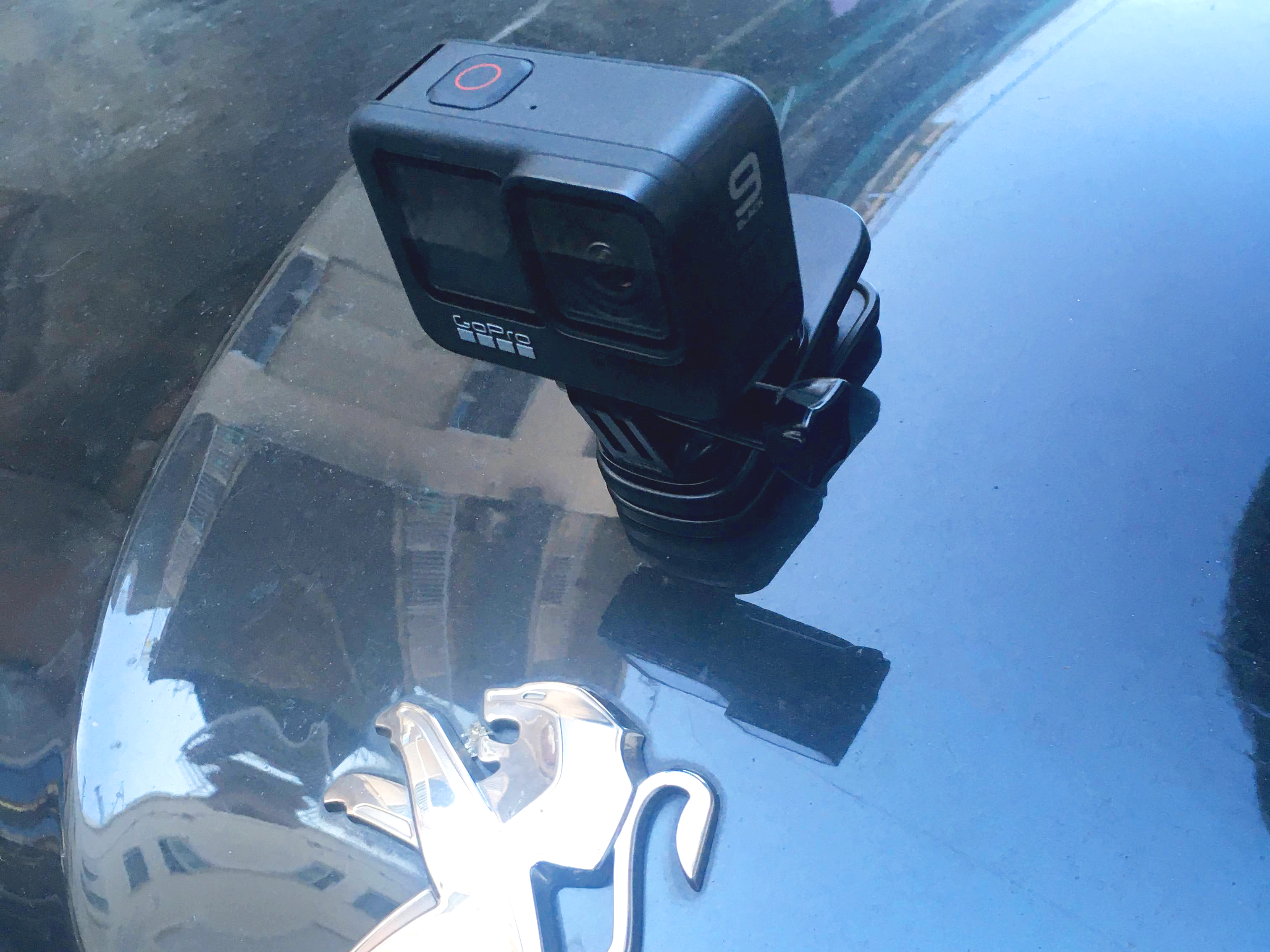}}
  \small\centerline{(a) Vehicle-mounted sensor}
\end{minipage}
\hfill
\begin{minipage}[b]{0.48\linewidth}
  \centering
  \centerline{\includegraphics[width=4.0cm]{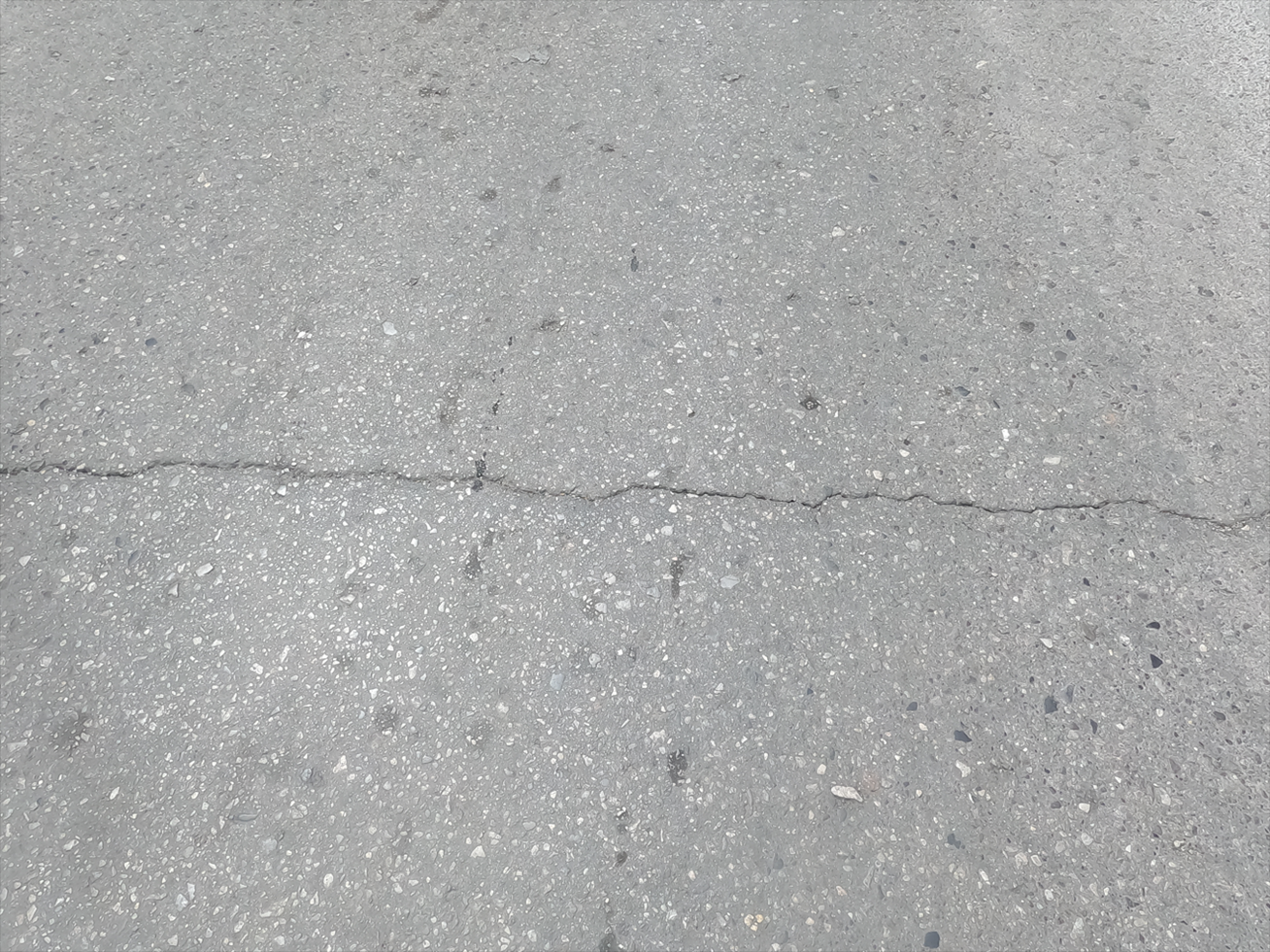}}
  \small\centerline{(b) Acquired RGB data}
\end{minipage}
\caption{Experimental setup for acquiring RGB road images.}
\label{fig:car}
\end{figure}

For the data collection process of the CrackMap dataset that constitutes the test dataset for the current study, a GoPro HERO9 Black was used. During the data acquisition process, the optical sensor was mounted on an inspection vehicle (see Fig. \ref{fig:car}a). It is emphasized that the acquired data are RGB images with an aspect ratio of 4:3 and, in particular, with a pixel resolution of 5,184×3,888 (see Fig. \ref{fig:car}b). Moreover, the RGB sensor was set to shoot at a high frame rate and, more specifically, at 50 frames per second in order to ensure sufficient data acquisition regarding both the positive (road surface with cracks) and negative (non-deteriorated road surface) events.

It is also highlighted that in order to deal with the severe class imbalance problem from the acquired RGB data, image patches with a resolution of 256×256 were manually extracted and then annotated. The patches were segmented and verified by engineer experts, within the framework of the H2020 HERON project \cite{10.1145/3529190.3534746}. As presented in Table \ref{table:data}, the CrackMap dataset contains 120 annotated images with a resolution of 256×256. We evaluate the comparative models that perform the crack segmentation task on the CrackMap data, minus the 6 extracted images that correspond to the 5\% of the lower ranked images and were eventually utilized for the refinement process (see Section \ref{sec:online}). CrackMap has been made available to the scientific community, for verifying the results and further research, at \href{https://github.com/ikatsamenis/CrackMap}{https://github.com/ikatsamenis/CrackMap}.

\subsection{Comparative algorithms and training configuration}
\label{sec:algorithms}

The validation of the proposed methodology for the crack segmentation task is based on examining its performance against other state-of-the-art approaches that perform crack recognition and precise localization in a different way. In particular, we compare the proposed static and dynamically refined R2AU-Net models on the CrackMap dataset (minus the 6 extracted images) with the following segmentation algorithms: (i) U-Net \cite{10.1007/978-3-319-24574-4_28}, (ii) V-Net \cite{milletari2016v}, (iii) ResU-Net \cite{8309343}, (iv) R2U-Net \cite{alom2018recurrent}, (v) Attention U-Net \cite{oktay2018attention}, and (vi) ResUNet-a \cite{DIAKOGIANNIS202094}.

The aforementioned deep models were developed using Keras and TensorFlow libraries in Python. In parallel, they were trained and evaluated on NVIDIA Tesla T4 GPU provided by Google Colab. We trained the neural networks for 100 epochs with early stopping criteria set to 10 epochs in order to avoid overfitting, using mini-batches of size 8. The training processes started from scratch by randomly initializing the networks' weights. The models are trained end-to-end using the Adam algorithm to optimize the dice loss function, in order to deal with class imbalance problems.
It is noted in parallel that the optimizer is set to its default parameters ($\beta_1=0.9$ and $\beta_2=0.999$), with an initial learning rate of $10^{-3}$ that is decayed by a factor of 10 each time there was no improvement in the validation loss, for 5 consecutive epochs. Lastly, for the rectification mechanism, we fine-tune the proposed R2AU-Net by retraining it for 5 epochs, with a learning rate reduced by a factor of 10, to avoid damaging its weights.

\begin{figure*}[ht]
\begin{minipage}[b]{0.106\linewidth}
  \centering
  \centerline{\includegraphics[width=1.95cm]{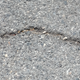}}
\end{minipage}
\hfill
\begin{minipage}[b]{0.106\linewidth}
  \centering
  \centerline{\includegraphics[width=1.95cm]{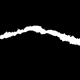}}
\end{minipage}
\hfill
\begin{minipage}[b]{0.106\linewidth}
  \centering
  \centerline{\includegraphics[width=1.95cm]{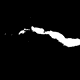}}
\end{minipage}
\hfill
\begin{minipage}[b]{0.106\linewidth}
  \centering
  \centerline{\includegraphics[width=1.95cm]{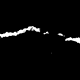}}
\end{minipage}
\hfill
\begin{minipage}[b]{0.106\linewidth}
  \centering
  \centerline{\includegraphics[width=1.95cm]{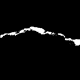}}
\end{minipage}
\hfill
\begin{minipage}[b]{0.106\linewidth}
  \centering
  \centerline{\includegraphics[width=1.95cm]{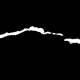}}
\end{minipage}
\hfill
\begin{minipage}[b]{0.106\linewidth}
  \centering
  \centerline{\includegraphics[width=1.95cm]{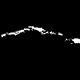}}
\end{minipage}
\hfill
\begin{minipage}[b]{0.106\linewidth}
  \centering
  \centerline{\includegraphics[width=1.95cm]{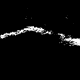}}
\end{minipage}
\hfill
\begin{minipage}[b]{0.106\linewidth}
  \centering
  \centerline{\includegraphics[width=1.95cm]{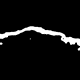}}
\end{minipage}
\hfill
%
\begin{minipage}[b]{0.106\linewidth}
  \centering
  \centerline{\includegraphics[width=1.95cm]{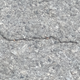}}
\end{minipage}
\hfill
\begin{minipage}[b]{0.106\linewidth}
  \centering
  \centerline{\includegraphics[width=1.95cm]{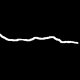}}
\end{minipage}
\hfill
\begin{minipage}[b]{0.106\linewidth}
  \centering
  \centerline{\includegraphics[width=1.95cm]{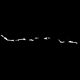}}
\end{minipage}
\hfill
\begin{minipage}[b]{0.106\linewidth}
  \centering
  \centerline{\includegraphics[width=1.95cm]{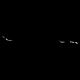}}
\end{minipage}
\hfill
\begin{minipage}[b]{0.106\linewidth}
  \centering
  \centerline{\includegraphics[width=1.95cm]{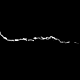}}
\end{minipage}
\hfill
\begin{minipage}[b]{0.106\linewidth}
  \centering
  \centerline{\includegraphics[width=1.95cm]{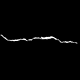}}
\end{minipage}
\hfill
\begin{minipage}[b]{0.106\linewidth}
  \centering
  \centerline{\includegraphics[width=1.95cm]{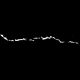}}
\end{minipage}
\hfill
\begin{minipage}[b]{0.106\linewidth}
  \centering
  \centerline{\includegraphics[width=1.95cm]{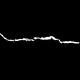}}
\end{minipage}
\hfill
\begin{minipage}[b]{0.106\linewidth}
  \centering
  \centerline{\includegraphics[width=1.95cm]{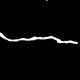}}
\end{minipage}
\hfill
%
\begin{minipage}[b]{0.106\linewidth}
  \centering
  \centerline{\includegraphics[width=1.95cm]{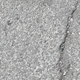}}
\end{minipage}
\hfill
\begin{minipage}[b]{0.106\linewidth}
  \centering
  \centerline{\includegraphics[width=1.95cm]{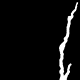}}
\end{minipage}
\hfill
\begin{minipage}[b]{0.106\linewidth}
  \centering
  \centerline{\includegraphics[width=1.95cm]{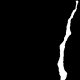}}
\end{minipage}
\hfill
\begin{minipage}[b]{0.106\linewidth}
  \centering
  \centerline{\includegraphics[width=1.95cm]{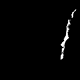}}
\end{minipage}
\hfill
\begin{minipage}[b]{0.106\linewidth}
  \centering
  \centerline{\includegraphics[width=1.95cm]{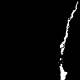}}
\end{minipage}
\hfill
\begin{minipage}[b]{0.106\linewidth}
  \centering
  \centerline{\includegraphics[width=1.95cm]{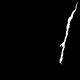}}
\end{minipage}
\hfill
\begin{minipage}[b]{0.106\linewidth}
  \centering
  \centerline{\includegraphics[width=1.95cm]{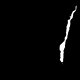}}
\end{minipage}
\hfill
\begin{minipage}[b]{0.106\linewidth}
  \centering
  \centerline{\includegraphics[width=1.95cm]{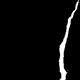}}
\end{minipage}
\hfill
\begin{minipage}[b]{0.106\linewidth}
  \centering
  \centerline{\includegraphics[width=1.95cm]{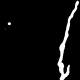}}
\end{minipage}
\hfill
%
\begin{minipage}[b]{0.106\linewidth}
  \centering
  \centerline{\includegraphics[width=1.95cm]{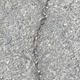}}
  \small\centerline{Input image}
\end{minipage}
\hfill
\begin{minipage}[b]{0.106\linewidth}
  \centering
  \centerline{\includegraphics[width=1.95cm]{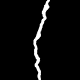}}
 \small\centerline{Ground truth}
\end{minipage}
\hfill
\begin{minipage}[b]{0.106\linewidth}
  \centering
  \centerline{\includegraphics[width=1.95cm]{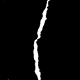}}
  \small\centerline{U-Net \cite{10.1007/978-3-319-24574-4_28}}
\end{minipage}
\hfill
\begin{minipage}[b]{0.106\linewidth}
  \centering
  \centerline{\includegraphics[width=1.95cm]{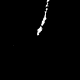}}
  \small\centerline{V-Net \cite{milletari2016v}}
\end{minipage}
\hfill
\begin{minipage}[b]{0.106\linewidth}
  \centering
  \centerline{\includegraphics[width=1.95cm]{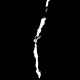}}
  \small\centerline{ResU-Net \cite{8309343}}
\end{minipage}
\hfill
\begin{minipage}[b]{0.106\linewidth}
  \centering
  \centerline{\includegraphics[width=1.95cm]{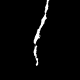}}
  \small\centerline{R2U-Net \cite{alom2018recurrent}}
\end{minipage}
\hfill
\begin{minipage}[b]{0.106\linewidth}
  \centering
  \centerline{\includegraphics[width=1.95cm]{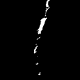}}
 \small\centerline{AttU-Net \cite{oktay2018attention}}
\end{minipage}
\hfill
\begin{minipage}[b]{0.106\linewidth}
  \centering
  \centerline{\includegraphics[width=1.95cm]{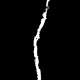}}
  \small\centerline{ResUNet-a \cite{DIAKOGIANNIS202094}}
\end{minipage}
\hfill
\begin{minipage}[b]{0.106\linewidth}
  \centering
  \centerline{\includegraphics[width=1.95cm]{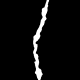}}
  \small\centerline{R2AU-Net}
\end{minipage}
%
\caption{Visual comparison of the static deep models' segmentation outputs.}
\label{fig:deeplearning}
\end{figure*}

\subsection{Experiments and comparisons}
\label{sec:comparisons}

Fig. \ref{fig:deeplearning} depicts a visual comparison of the output masks generated by our method and the aforementioned comparative models. For a quantitative analysis of the experimental results, the performance of the implemented models is evaluated in terms of the Dice coefficient and IoU metric. In particular, we compute the aforementioned metrics for every RGB image of the CrackMap dataset (minus the 6 extracted images) and, thereby, we report the average values across all 114 images of the test set with a confidence level of 95\%. As can be observed in Table \ref{table:test} the proposed R2AU-Net outperforms the various state-of-the-art algorithms by at least 1.69\% and 1.89\% in terms of Dice and IoU scores respectively.

\begin{table}[h!]
\small
\centering
\begin{tabular}{||c c c||}
 \hline
 \textbf{Model} & \textbf{Avg. Dice} & \textbf{Avg. IoU} \\ [0.5ex] 
 \hline\hline
 U-Net \cite{10.1007/978-3-319-24574-4_28} & 49.73\% $\pm$ 3.75\% & 35.54\% $\pm$ 3.36\% \\ 
 V-Net \cite{milletari2016v} & 38.04\% $\pm$ 4.65\% & 26.57\% $\pm$ 3.68\% \\
 ResU-Net \cite{8309343} & 60.37\% $\pm$ 2.50\% & 44.53\% $\pm$ 2.48\% \\
 R2U-Net \cite{alom2018recurrent} & 70.74\% $\pm$ 1.59\% & 55.39\% $\pm$ 1.84\% \\
 AttU-Net \cite{oktay2018attention} & 52.64\% $\pm$ 3.70\% & 38.06\% $\pm$ 3.21\% \\
 ResUNet-a \cite{DIAKOGIANNIS202094} & 63.28\% $\pm$ 2.47\% & 47.60\% $\pm$ 2.49\% \\ 
 \textbf{R2AU-Net} & \textbf{72.43\% $\pm$ 1.36\%} & \textbf{57.28\% $\pm$ 1.61\%} \\
 \textbf{FS R2AU-Net} & \textbf{77.06\% $\pm$ 1.17\%} & \textbf{63.11\% $\pm$ 1.50\%} \\ [1ex] 
 \hline
\end{tabular}
\caption{Performance evaluation and comparisons.}
\label{table:test}
\end{table}

\begin{figure}[htb]

\begin{minipage}[b]{.24\linewidth}
  \centering
  \centerline{\includegraphics[width=2.1cm]{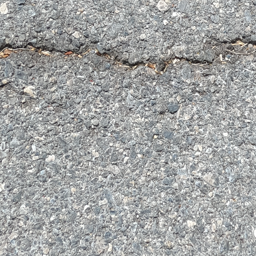}}
\end{minipage}
\hfill
\begin{minipage}[b]{0.24\linewidth}
  \centering
  \centerline{\includegraphics[width=2.1cm]{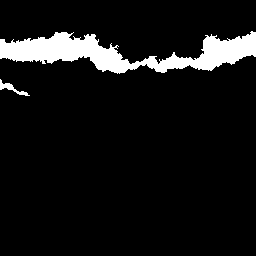}}
\end{minipage}
\begin{minipage}[b]{.24\linewidth}
  \centering
  \centerline{\includegraphics[width=2.1cm]{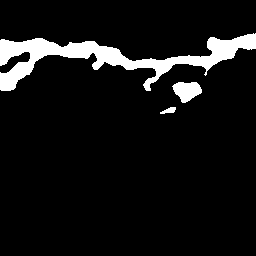}}
\end{minipage}
\hfill
\begin{minipage}[b]{0.24\linewidth}
  \centering
  \centerline{\includegraphics[width=2.1cm]{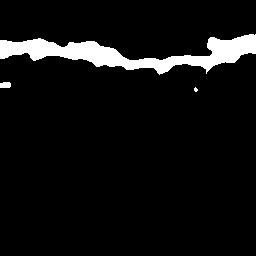}}
\end{minipage}
\begin{minipage}[b]{.24\linewidth}
  \centering
  \centerline{\includegraphics[width=2.1cm]{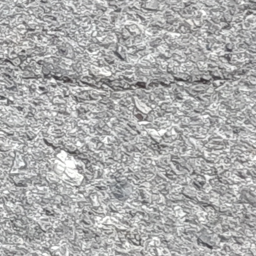}}
\end{minipage}
\hfill
\begin{minipage}[b]{0.24\linewidth}
  \centering
  \centerline{\includegraphics[width=2.1cm]{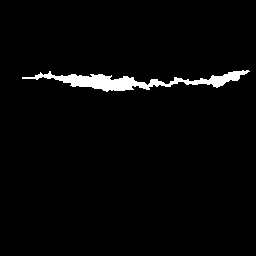}}
\end{minipage}
\begin{minipage}[b]{.24\linewidth}
  \centering
  \centerline{\includegraphics[width=2.1cm]{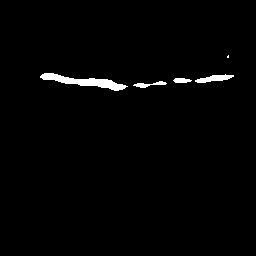}}
\end{minipage}
\hfill
\begin{minipage}[b]{0.24\linewidth}
  \centering
  \centerline{\includegraphics[width=2.1cm]{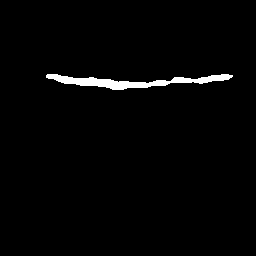}}
\end{minipage}
%
%
\begin{minipage}[b]{.24\linewidth}
  \centering
  \centerline{\includegraphics[width=2.1cm]{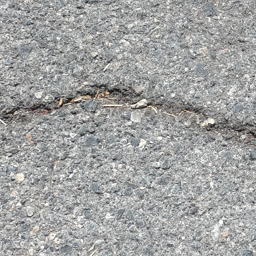}}
  \small\centerline{Input image}
\end{minipage}
\hfill
\begin{minipage}[b]{0.24\linewidth}
  \centering
  \centerline{\includegraphics[width=2.1cm]{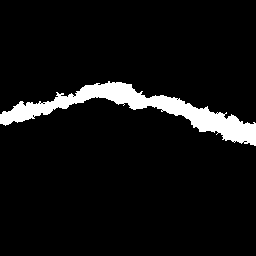}}
  \small\centerline{Ground truth}
\end{minipage}
\begin{minipage}[b]{.24\linewidth}
  \centering
  \centerline{\includegraphics[width=2.1cm]{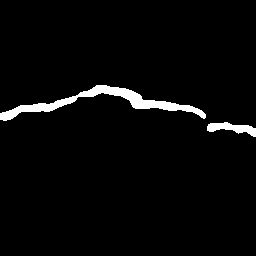}}
  \small\centerline{Original}
\end{minipage}
\hfill
\begin{minipage}[b]{0.24\linewidth}
  \centering
  \centerline{\includegraphics[width=2.1cm]{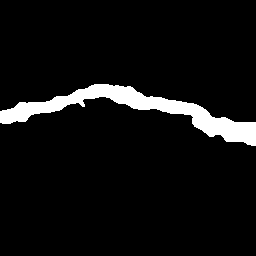}}
  \small\centerline{Rectified}
\end{minipage}
\caption{Output masks produced by the original R2AU-Net against the proposed rectified few-shot learning paradigm.}
\label{fig:staticVSonline}
\end{figure}

Finally, Fig. \ref{fig:staticVSonline} illustrates indicative segmentation outputs of the R2AU-Net before and after applying the proposed few-shot rectification mechanism. As shown in Table \ref{table:test}, the rectified R2AU-Net model demonstrated increased performance of 4.63\% and 5.83\% in terms of Dice and IoU respectively, after the few-shot refinement procedure. To investigate whether this improvement is statistically significant, we exploit the Wilcoxon signed-rank test on the obtained scores of the two models, which is a nonparametric statistical test that compares two paired groups \cite{wilcoxon1992individual}. The obtained {\it p}-values for both metrics are lower than .001 and, thus, we can reject the null hypothesis, which entails that there is a statistically significant difference in the comparative results of the R2AU-Net, before and after the proposed few-shot refinement process.

\section{CONCLUSION}
\label{sec:refs}


This paper presents a few-shot learning strategy for road crack segmentation. The scheme is based on R2AU-Net which exploits recurrent residual and attention mechanisms to capture richer global context information and local semantic features. Also, the adopted few-shot refinement process, through which the network weights are dynamically updated as a few incoming rectified samples are fed into the algorithm, led to state-of-the-art performance on a new publicly available dataset.

\bibliographystyle{IEEEbib}
\bibliography{strings,refs}

\end{document}